\pdfoutput=1
\documentclass[11pt]{article}

\usepackage{acl}

\usepackage{times}
\usepackage{latexsym}

\usepackage[T1]{fontenc}
\usepackage[utf8]{inputenc}

\usepackage{microtype}

\usepackage{inconsolata}

\usepackage{color}

\usepackage{booktabs}
\usepackage{multirow}
\usepackage{graphicx}
\usepackage{colortbl}
\usepackage{tikz}
\usepackage{forest}
\usetikzlibrary{trees,positioning,shapes,shadows,arrows.meta}

\definecolor{indigo}{HTML}{4B0082}
\definecolor{amber}{HTML}{cc4e00}
\definecolor{teal}{HTML}{008080}
\definecolor{grey}{HTML}{979ea8}

\title{Can Knowledge Graphs Reduce Hallucinations in LLMs? : A Survey}
\author{Garima Agrawal ~~ Tharindu Kumarage ~~ Zeyad Alghamdi ~~ Huan Liu \\
         Arizona State University \\
        \texttt{\{garima.agrawal, kskumara, zalgham1, huanliu\}@asu.edu}}

\begin{document}
\maketitle
\begin{abstract}
The contemporary LLMs are prone to producing hallucinations, stemming mainly from the knowledge gaps within the models. To address this critical limitation, researchers employ diverse strategies to augment the LLMs by incorporating external knowledge, aiming to reduce hallucinations and enhance reasoning accuracy. Among these strategies, leveraging knowledge graphs as a source of external information has demonstrated promising results. In this survey, we comprehensively review these knowledge-graph-based augmentation techniques in LLMs, focusing on their efficacy in mitigating hallucinations. We systematically categorize these methods into three overarching groups, offering methodological comparisons and performance evaluations. Lastly, this survey explores the current trends and challenges associated with these techniques and outlines potential avenues for future research in this emerging field.
\end{abstract}

%The large language models (LLMs) exhibit remarkable reasoning and natural language generation capabilities. 

% degree to which mitigation
% These instructions are for authors submitting papers to *ACL conferences using \LaTeX. They are not self-contained. All authors must follow the general instructions for *ACL proceedings,\footnote{\url{http://acl-org.github.io/ACLPUB/formatting.html}} and this document contains additional instructions for the \LaTeX{} style files.

% The templates include the \LaTeX{} source of this document (\texttt{acl.tex}),
% the \LaTeX{} style file used to format it (\texttt{acl.sty}),
% an ACL bibliography style (\texttt{acl\_natbib.bst}),
% an example bibliography (\texttt{custom.bib}),
% and the bibliography for the ACL Anthology (\texttt{anthology.bib}).
\section{Introduction}
\label{intro}
 
 Large language models (LLMs) seek to emulate human intelligence through statistical training on extensive datasets~\cite{huang2022towards}. LLMs operate on input text to predict the subsequent token or word in the sequence while identifying patterns and connections between words and phrases, aiming to comprehend and generate human-like text. Due to their stochastic decoding processes, i.e., sampling the next token in the sequence, these models exhibit probabilistic behavior, potentially yielding varied outputs or predictions for the same input across different instances. Additionally, if the training data includes misinformation, biases, or inaccuracies, these flaws may be mirrored or amplified in the content produced by these models. LLMs also face challenges in accurately interpreting phrases or terms when the context is vague and resides in a knowledge gap region of the model, leading to outputs that may sound plausible but are often irrelevant or incorrect~\cite{ji2023survey, lenat2023getting}. This phenomenon, often termed "hallucinations," undermines the reliability of these models~\cite{mallen2023not}.

 \begin{figure}
    \centering
    \includegraphics[width=0.9\linewidth]{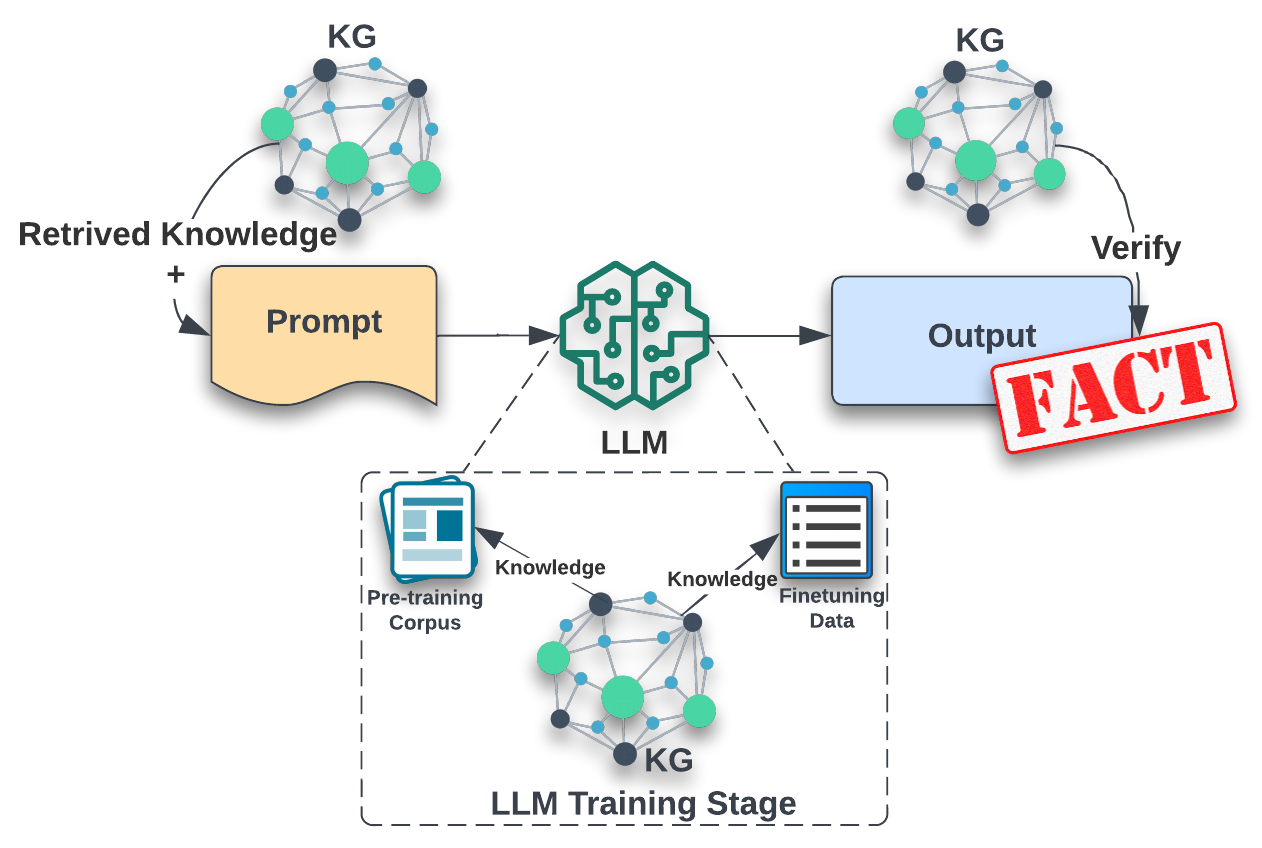}
    \caption{Knowledge Graphs (KG) employed to reduce hallucinations in LLMs at different stages.}
    \label{fig:intro}
\end{figure}

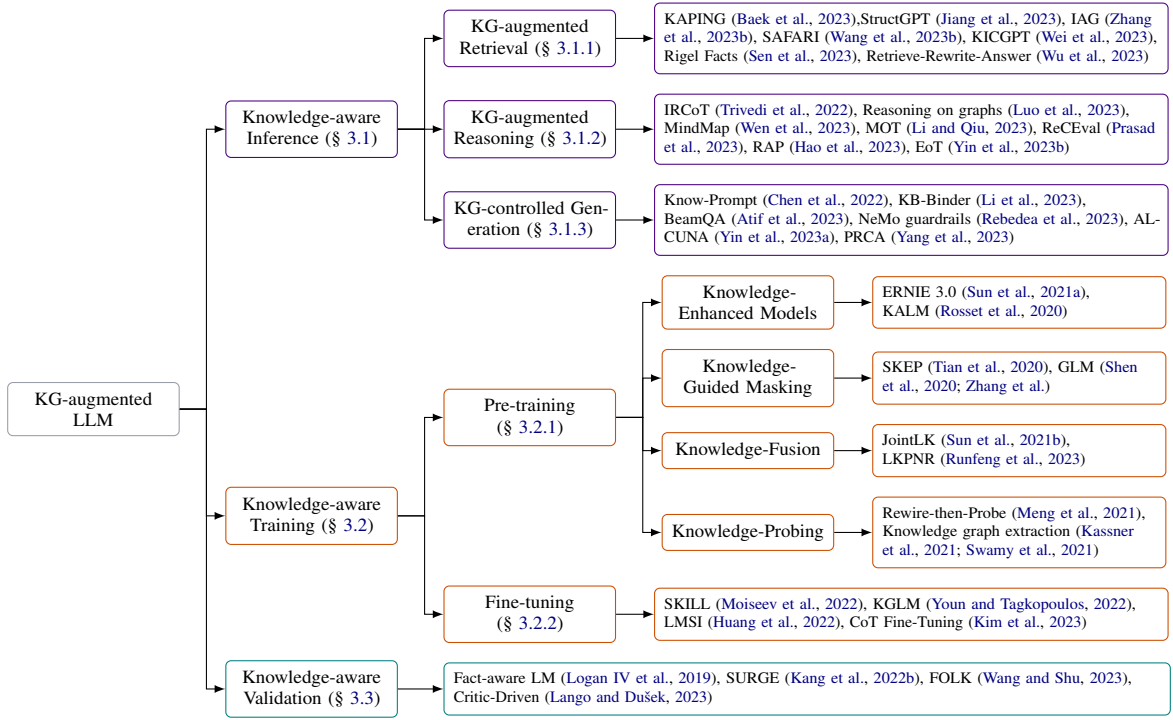
\begin{figure*}
    \centering

\tikzset{
    basic/.style  = {draw, text width=2cm, align=center, rectangle, font=\scriptsize},
    root/.style   = {basic, draw=grey, rounded corners=2pt, thin, align=center, fill=none},
    infnode/.style = {basic, draw=indigo, thin, rounded corners=2pt, align=left, fill=none, text width=2cm, align=center},
    tranode/.style = {basic,draw=amber, thin, rounded corners=2pt, align=center, fill=none,text width=2cm},
    valnode/.style = {basic, draw=teal, thin, rounded corners=2pt, align=center, fill=none, text width=2cm},
    infcitenode/.style = {infnode, thin, align=left, fill=none, text width=65mm, font=\tiny},
    tracitenode/.style = {tranode, thin, align=left, fill=none, text width=65mm, font=\tiny},
    pretracitenode/.style = {tranode, thin, align=left, fill=none, text width=36mm, font=\tiny},
    valcitenode/.style = {valnode, thin, align=left, fill=none, text width=93mm, font=\tiny},
    edge from parent/.style={draw=black, edge from parent fork right}

}
\begin{forest} for tree={
    grow=east,
    growth parent anchor=west,
    parent anchor=east,
    child anchor=west,
    anchor=center,
    edge path={\noexpand\path[\forestoption{edge},->, >={latex}] 
         (!u.parent anchor) -- +(10pt,0pt) |-  (.child anchor) 
         \forestoption{edge label};}
}
% lsep is used for arrow distance
[KG-augmented LLM, root,  l sep=6mm,
    [Knowledge-aware Validation (\S~\ref{sec:KA_valid}), valnode,  l sep=6mm,
        [{Fact-aware LM~\cite{logan2019barack}, SURGE~\cite{kang2022knowledge}, FOLK~\cite{wang2023explainable}, Critic-Driven~\cite{lango2023critic}}, valcitenode
        ]
    ]
    [Knowledge-aware Training (\S~\ref{sec:KA_train}), tranode,  l sep=6mm,
        [Fine-tuning (\S~\ref{train:ft}), tranode,l sep=5mm,
            [{SKILL~\cite{moiseev2022skill},  KGLM~\cite{youn2022kglm}, LMSI~\cite{huang2022large}, CoT Fine-Tuning~\cite{kim2023cot}}, tracitenode
            ]
        ] 
        [Pre-training (\S~\ref{train:pre}), tranode,l sep=6mm,
            [Knowledge-Probing, tranode,l sep=5mm,
            [{Rewire-then-Probe~\cite{meng2021rewire}, Knowledge graph extraction~\cite{kassner2021multilingual,swamy2021interpreting}},pretracitenode]
            ] 
            [Knowledge-Fusion, tranode,l sep=5mm,
            [{JointLK~\cite{sun2021jointlk},  LKPNR~\cite{runfeng2023lkpnr}},pretracitenode]
            ] 
            [Knowledge-Guided Masking, tranode,l sep=5mm,
            [{SKEP~\cite{tian2020skep}, GLM~\cite{shen2020exploiting,zhangbert}},pretracitenode]
            ] 
            [Knowledge-Enhanced Models, tranode,l sep=5mm,
            [{ERNIE 3.0~\cite{sun2021ernie}, KALM~\cite{rosset2020knowledge}},
            pretracitenode]
            ] 
        ] 
    ]
    [Knowledge-aware Inference (\S~\ref{sec:KA_infer}), infnode,  l sep=6mm,
        [KG-controlled Generation (\S~\ref{infer:ctrl}), infnode,l sep=5mm,
            [{Know-Prompt~\cite{chen2022knowprompt}, KB-Binder~\cite{li2023few}, BeamQA~\cite{atif2023beamqa}, NeMo guardrails~\cite{rebedea2023nemo}, ALCUNA~\cite{yin2023alcuna}, PRCA~\cite{yang2023prca}}, infcitenode]
        ]
        [KG-augmented Reasoning (\S~\ref{infer:reas}), infnode,l sep=5mm,
            [{IRCoT~\cite{trivedi2022interleaving}, Reasoning on graphs~\cite{luo2023reasoning}, MindMap~\cite{wen2023mindmap}, MOT~\cite{li2023mot}, ReCEval~\cite{prasad2023receval}, RAP~\cite{hao2023reasoning}, EoT~\cite{yin2023exchange}}, infcitenode]]
        [KG-augmented Retrieval (\S~\ref{infer:ret}), infnode,l sep=5mm,            [{KAPING~\cite{baek2023knowledge},StructGPT~\cite{jiang2023structgpt}, IAG~\cite{zhang2023iag}, SAFARI~\cite{wang2023large}, KICGPT~\cite{wei2023kicgpt}, Rigel Facts~\cite{sen2023knowledge}, Retrieve-Rewrite-Answer~\cite{wu2023retrieve}}, infcitenode]]
         ] 
    ]
\end{forest}
    \caption{Taxonomy of Knowledge Graph-Augmented Large Language Models}
    \label{fig_tax}
\end{figure*}

Addressing the issue of hallucinations in these models is challenging due to their inherent probabilistic nature.
To effectively tackle this issue, there have been continuous research efforts in making knowledge updates and model tuning~\cite{zhang2023large, mialon2023augmented, petroni2019language}.
However, adding random information does not improve the model's interpretation and reasoning capabilities. Instead, providing more granular and contextually relevant, precise external knowledge can significantly aid the model in recalling essential information~\cite{jiang2020can}.

One emerging research trend is enhancing LLMs through integrating knowledge representation tools such as knowledge graphs (KGs)~\cite{mruthyunjaya2023rethinking}. Zheng et al.~\cite{zheng2023does} demonstrate that augmenting these models with comprehensive external knowledge from KGs can boost their performance and facilitate a more robust reasoning process. 
The strategies for enhancing LLMs with KGs can be grouped into three main categories, each uniquely contributing to the refinement of the model as shown in 
Figure~\ref{fig:intro}: enhancing the inference process, improving the learning mechanism, and establishing robust methods for validating the model's decisions. 
% We coin these categories as \textit{Knowledge-Aware Inference}, \textit{Knowledge-Aware Learning}, and \textit{Knowledge-Aware Validation}. 

In this survey, we critically review KG augmentation methods used in specific stages to reduce hallucinations in LLMs and improve their performance and reliability. In Section~\ref{method}, we classify these methods into three overarching categories: \textbf{(1) \textit{Knowledge-Aware Inference}}, \textbf{(2) \textit{Knowledge-Aware Learning}}, and \textbf{(3) \textit{Knowledge-Aware Validation}}. Additionally, in Section~\ref{discussion}, we evaluate the empirical efficacy of these methods and discuss current research trends, followed by suggestions for potential future research directions.

\noindent \textbf{Related Works:} There are several related surveys which discuss LLM augmentation using external knowledge~\cite{hu2023survey, yin2022survey, alkhamissi2022review, ye2022generative, wei2021knowledge, liang2022reasoning, zhang2023large, mialon2023augmented}. However, to our knowledge, this is the first survey to exclusively focus on critically reviewing LLM augmentation methods utilizing structured knowledge from knowledge graphs. Specifically, our emphasis is on addressing hallucinations in LLMs through KG integration.

\section{Preliminaries}
\label{prelim}
We now introduce the preliminaries and definitions that will be used throughout the survey.

\subsection{Large Language Models}
Language modeling, a key task in natural language processing (NLP), focuses on understanding language's structure and generating text. It has gained importance over recent years. Specifically, in neural probabilistic language models~\cite{bengio2000neural}, the goal is to estimate the likelihood of a text sequence. It involves computing the probability of each token $x_i$ in the sequence, considering preceding tokens, using the chain rule to simplify the process.

\begin{equation} 
p(x) = \prod_{i=1}^{N} p(x_i | x_1, x_1 ... x_{i-1}) 
\label{eq:lm} 
\end{equation}

The introduction of the transformer architecture~\cite{vaswani2017attention} significantly advanced neural probabilistic language models, enabling efficient parallel processing and recognition of long-range dependencies in text. Coupled with training advancements like instruction tuning and Reinforcement Learning from Human Feedback (RLHF)~\cite{ouyang2022training}, these neural probabilistic language models led to the creation of advanced Large Language Models (LLMs) like GPT-3~\cite{brown2020language}, GPT-4~\cite{openai2023gpt4}, and PaLM~\cite{chowdhery2022palm}, notable for their exceptional language capabilities.

\subsection{Knowledge Graphs}
Knowledge graphs (KGs) organize information into a structured format, capturing relationships between real-world entities, making it comprehensible to both humans and machines~\cite{hogan2021knowledge}. They store data as triples in a graph, with nodes representing entities (like people or places) and edges depicting relationships. Their capacity to represent complex interrelations makes them applicable in various domains~\cite{fensel2020we}. KGs are used in a semantic search to enhance search engines semantic understanding~\cite{singhal2012introducing}, enterprise knowledge management~\cite{deng2023construction}, supply chain optimization~\cite{deng2023research}, education~\cite{agrawal2022building}, financial fraud detection~\cite{mao2022financial}, cybersecurity~\cite{agrawal2023aiseckg}, recommendation systems~\cite{guo2020survey}, and QA systems~\cite{agrawal2023auction, omar2023universal, jiang2021research}.

\section{Knowledge Graph-Enhanced LLMs}
\label{method}

The LLMs primarily have three points of failure: a failure to comprehend the question due to lack of context, insufficient knowledge to respond accurately, or an inability to recall specific facts. Improving the cognitive capabilities of these models involves refining their inference-making process, optimizing learning mechanisms, and establishing a mechanism to validate results. 
This survey comprehensively reviews existing methodologies aimed at mitigating hallucinations and enhancing the reasoning capabilities of LLMs through the augmentation of KGs using these three techniques. We classify them as \textbf{\textit{Knowledge-Aware Inference}, \textit{Knowledge-Aware Learning}, and \textit{Knowledge-Aware Validation}}. Figure~\ref{fig_tax} details key works from each of these categories.

\begin{figure}[b]
    \centering
    \includegraphics[width=0.48\textwidth]{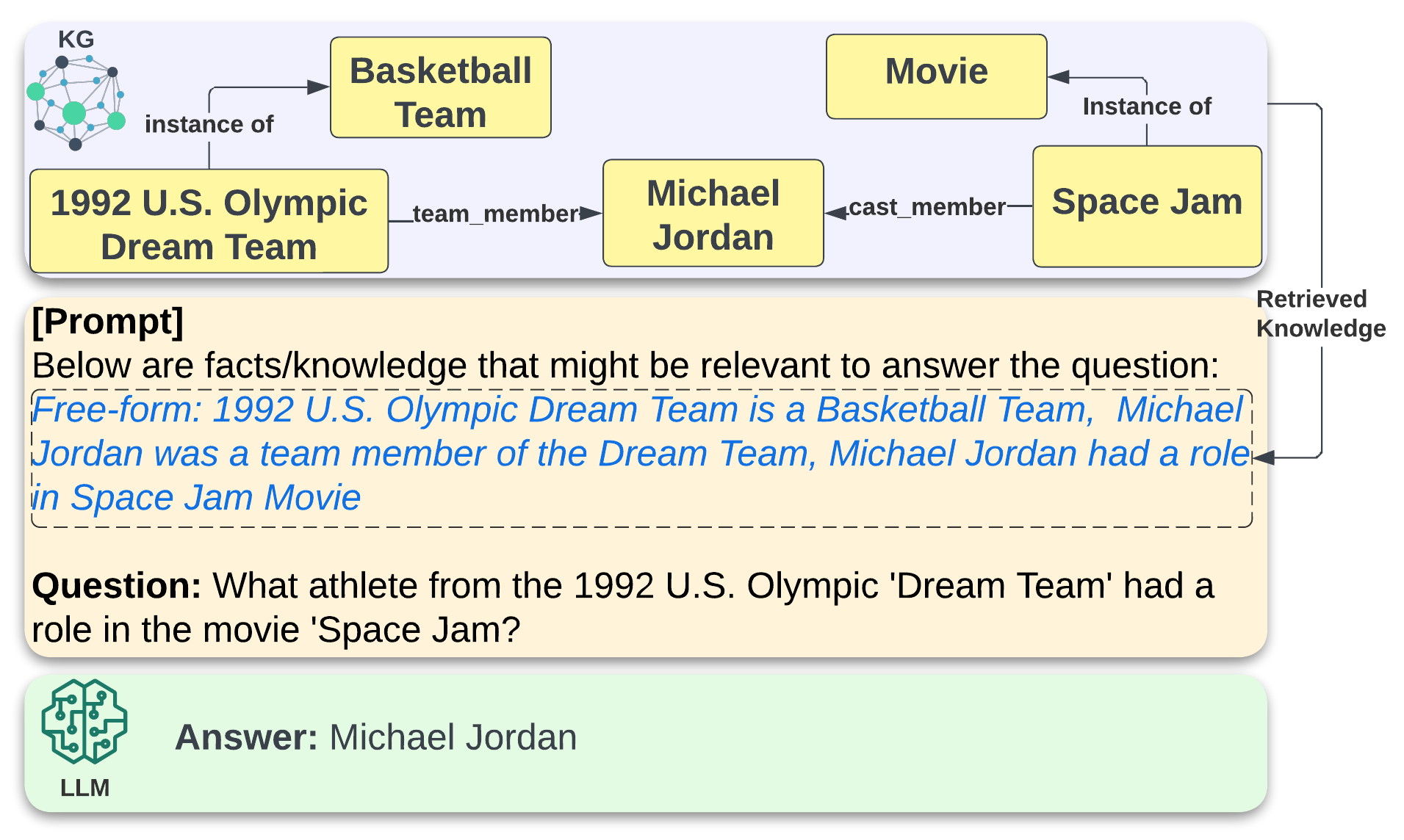}
    \caption{Knowledge-aware inference by incorporating KG-augmented retrieval~\cite{baek2023knowledge}.}
    \label{fig:kg_infer}
\end{figure}

\subsection{Knowledge-Aware Inference}
\label{sec:KA_infer}

In LLMs, \textit{``inference''} means generating text or predictions from a pre-trained model based on an input context. Challenges include incorrect or sub-optimal outputs due to ambiguous inputs, unclear context, knowledge gaps, training data biases, or inability to generalize to unseen scenarios. LLMs often struggle with multi-step reasoning and, unlike humans, can not seek extra information to clarify ambiguous queries.
To improve LLMs' inference and reasoning, researchers integrate KGs for structured symbolic knowledge, primarily by incorporating them at the input level to enhance contextual understanding. These methods, are further categorized into `KG-Augmented Retrieval,' `KG-Augmented Reasoning,' and `KG-Controlled Generation.'

\subsubsection{KG-Augmented Retrieval}
\label{infer:ret}

Retrieval-augmented generation models like RAG~\cite{lewis2020retrieval} and RALM~\cite{ram2023context} enhance LLMs' contextual awareness for knowledge-intensive tasks by providing relevant documents during generation, reducing hallucination without altering the LLM architecture. These methods, which are helpful for tasks needing external knowledge, augment 
top-k relevant documents to inputs. However, as shown in 
Figure~\ref{fig:kg_infer}, using well-organized, curated knowledge from structured sources or knowledge graphs, aligns more closely with factual accuracy.
Baek et al.~\cite{baek2023knowledge} introduced KAPING, which matches entities in questions to retrieve related triples from knowledge graphs for zero-shot question answering. Wu et al.~\cite{wu2023retrieve} found that converting these triples into textualized statements enhances LLM performance. 
Sen et al.~\cite{sen2023knowledge} developed a retriever module trained on a KGQA model, addressing the inadequacy of similarity-based retrieval for complex questions. StructGPT~\cite{jiang2023structgpt} augments LLMs with data from knowledge graphs, tables, and databases, utilizing structured queries for information extraction. Other notable works include IAG\cite{zhang2023iag}, KICGPT~\cite{wei2023kicgpt}, and SAFARI~\cite{wang2023large}.

LLMs serve as natural language interfaces, extracting and generating information without relying on their internal knowledge. Tools like the ChatGPT plugin use Langchain~\cite{langchain_2022} and LlamaIndex~\cite{Liu_LlamaIndex_2022} to integrate external data, prompting LLMs for context-retrieved, knowledge-augmented outputs. However, relying solely on internal databases can limit performance due to restricted knowledge bases.
Mallen et al.~\cite{mallen2023not} investigated LLMs' factual knowledge retention, finding that augmenting with retrieved data improves performance. However, these models perform well with popular entities and relations but face challenges with less popular subjects, and increasing model size doesn't improve their performance in such cases.

\subsubsection{KG-Augmented Reasoning}
\label{infer:reas}
KG-augmented retrieval methods effectively answer factual questions. However, questions that require reasoning call for more proficient approaches, such as decomposing complex, multi-step tasks into manageable sub-queries, as detailed by~\cite{qiao2022reasoning, liu2023pre}. These techniques are referred to as KG-augmented reasoning methods in our study.
Following the intuition behind the human reasoning process, the 
Chain of Thought (CoT)~\cite{wei2022chain}, 
Chain of Thought with Self-Consistency (CoT-SC)~\cite{wang2022self}, 
Program-Aided Language Model (PAL)~\cite{gao2023pal}, and 
Reason and Act (ReAct)~\cite{yao2022react}, 
Reflexion~\cite{shinn2023reflexion} methods used 
a series of intermediate reasoning steps to improve the complex reasoning ability of LLMs. These methods mimic human step-by-step reasoning, aiding in understanding and debugging the model's reasoning process. They are useful for math problems, commonsense reasoning, and symbolic tasks solvable through language-explained steps. Tree of Thoughts(ToT)~\cite{yao2023tree} method enhances this by exploring coherent text units as intermediate steps, enabling LLMs to consider multiple paths, self-evaluate, and make informed decisions. 

Different knowledge augmentation techniques using knowledge graphs, inspired by CoT and ToT prompting, enhance reasoning in domain-specific and open-domain tasks. ``Rethinking with Retrieval"~\cite{he2022rethinking} model uses decomposed reasoning steps from chain-of-thought prompting to retrieve external knowledge, leading to more accurate and faithful explanations. IRCoT~\cite{trivedi2022interleaving} interleaves generating chain-of-thoughts (CoT) and retrieving knowledge from graphs, iteratively guiding retrieval and reasoning for multi-step questions. MindMap~\cite{wen2023mindmap} introduces a plug-and-play approach to evoke graph-of-thoughts reasoning in LLMs. Reasoning on Graphs (RoG)~\cite{luo2023reasoning} uses knowledge graphs to create faithful reasoning paths based on various relations, enabling interpretable and accurate reasoning in LLMs.
Complementary advancements include MoT~\cite{li2023mot}, Democratizing Reasoning~\cite{wang2023democratizing}, ReCEval~\cite{prasad2023receval}, RAP~\cite{hao2023reasoning}, EoT~\cite{yin2023exchange} and Tree Prompting~\cite{singh2023tree}, each contributing uniquely to the development of reasoning capabilities in LLMs.

Exploring the interaction between prompts and large language models in the context of reasoning tasks is an exciting research avenue~\cite{liu2023pre}. A crucial aspect is the design of prompts tailored to the specific use case. However, the fundamental question of whether neural networks genuinely engage in "reasoning" remains unanswered, and it is uncertain whether following the correct reasoning path always leads to accurate answers ~\cite{qiao2022reasoning, jiang2020can}. 

\subsubsection{Knowledge-Controlled Generation}
\label{infer:ctrl}

These methods generate knowledge using a language model and then use probing or API calls for tasks. 
Liu et al.~\cite{liu2021generated} used a second model to produce question-related knowledge statements for deductions. 
Binder~\cite{cheng2022binding} uses Codex to parse context and generate task API calls. KB-Binder~\cite{li2023few} also employs Codex to create logical drafts for questions, integrating knowledge graphs for complete answers. Brate et al.~\cite{brate2022improving} create cloze-style prompts for entities in knowledge graphs, enhancing them with auxiliary data via SPARQL queries, improving recall and accuracy. KnowPrompt~\cite{chen2022knowprompt} generates prompts from a pre-trained model and tunes them for relation extraction in cloze-style tasks. BeamQA~\cite{atif2023beamqa} uses a language model to generate inference paths for knowledge graph embedding-based search in link prediction. ALCUNA~\cite{yin2023alcuna} and PRCA~\cite{yang2023prca} are other significant methods in controlled generation.

Guardrails in generative AI set operational boundaries for models, ensuring safe and secure output generation. NeMo guardrails~\cite{rebedea2023nemo} by Nvidia guide conversational flows in enterprise applications to meet safety and security standards. Knowledge-controlled generation ensures alignment with facts and prevents misinformation. Knowledge graph ontologies can provide specific domain constraints, aiding LLMs in defining output generation boundaries.

\subsection{Knowledge-Aware Training}
\label{sec:KA_train}

Another stage where we can address hallucination issues in LLMs is to utilize KGs to optimize their learning either by improving the quality of training data at the model pre-training stage or by fine-tuning the pre-trained language model (PLM) to adapt to specific tasks or domains. We classify these methods as \textit{Knowledge-Aware Pre-Training} and \textit{Knowledge-Aware Fine-Tuning}.

\subsubsection{Knowledge-Aware Pre-Training}
\label{train:pre}
\begin{figure}
    \centering
    \includegraphics[width=0.48\textwidth]{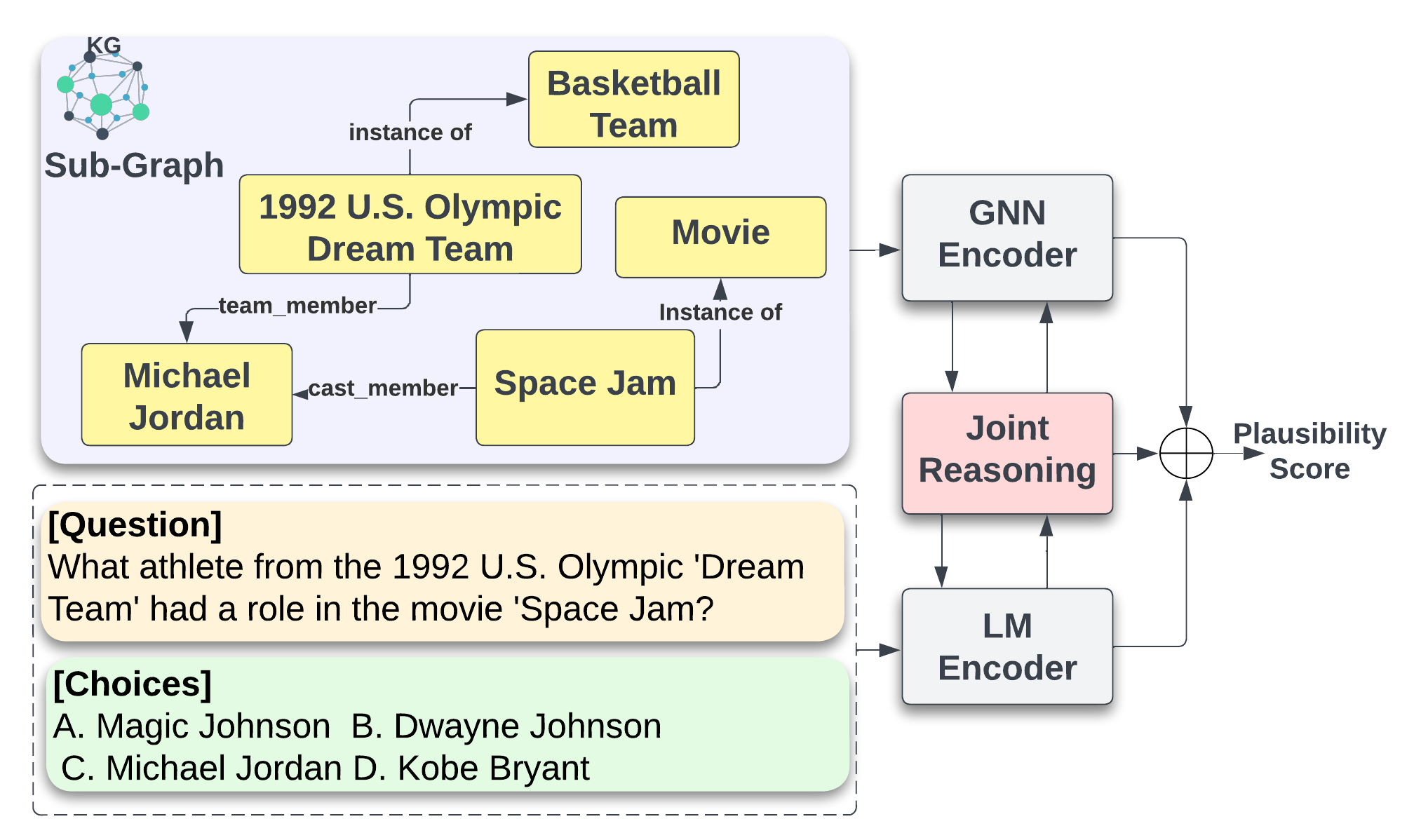}
    \caption{Knowledge-aware Pre-training by Knowledge Fusion ~\cite{sun2021jointlk}.}
    \label{fig:kg_train}
\end{figure}

Training data quality and diversity are crucial for reducing hallucinations in LLMs. Integrating knowledge graphs, which provide structured information about entities and their interconnections, improves the comprehension abilities of LLMs and aids in generating text that more accurately reflects the complexities of real-world entities. However, training from scratch is highly resource-heavy and expensive. Different approaches were proposed by researchers~\cite{yu2023pre, fu2023revisiting, deng2023construction, liu2020k, poerner2019bert, peters2019knowledge} for pre-training models by augmenting knowledge graphs in training data. We further categorize them as follows:

\begin{enumerate}
    \item \textit{\textbf{Knowledge-Enhanced Models}}: 
    These methods enriched the large-scale text corpora with KGs for improved language representation. ERNIE~\cite{zhang2019ernie} used masked language modeling (MLM) and next sentence prediction (NSP) in pre-training to capture the text's lexical and syntactical elements, combining context with knowledge facts for predictions. ERNIE 3.0~\cite{sun2021ernie} further evolved by integrating an auto-regressive model with an auto-encoding network, addressing the limitations of a single auto-regressive framework in exploring enhanced knowledge. Meanwhile, Rosset et al.~\cite{rosset2020knowledge} introduced a knowledge-aware input through an entity tokenizer dictionary, enhancing semantic understanding without altering the transformer architecture.
    
    \item \textit{\textbf{Knowledge-Guided Masking}}: Knowledge graph-guided entity masking schemes~\cite{shen2020exploiting,zhangbert} utilized linked knowledge graphs to mask key entities in texts, enhancing question-answering and knowledge-base completion tasks by leveraging relational knowledge. Similarly, Sentiment Knowledge Enhanced Pre-training (SKEP)~\cite{tian2020skep} employed sentiment masking to develop unified sentiment representations, improving performance across various sentiment analysis tasks.
    
    \item \textit{\textbf{Knowledge-Fusion}}: These methods integrates the KGs into LLMs using graph query encoders~\cite{wang2021kepler, ke2021jointgt, he2019integrating}. As shown in Figure~\ref{fig:kg_train}, JointLK~\cite{sun2021jointlk} employed knowledge fusion and joint reasoning for commonsense question answering, selectively using relevant KG nodes and synchronizing updates between text and graph encoders. LKPNR~\cite{runfeng2023lkpnr} combined LLMs with KGs, enhancing semantic understanding in complex news texts to create a personalized news recommendation framework through a KG-augmented encoder.

    \item \textit{\textbf{Knowledge-Probing}}: Knowledge probing involves examining language models to assess their factual and commonsense knowledge~\cite{petroni2019language}. This process aids in evaluating and enhancing the models~\cite{kassner2021multilingual,swamy2021interpreting}. Rewire-then-Probe~\cite{meng2021rewire} introduced a self-supervised contrastive-probing approach, utilizing biomedical knowledge graphs to learn language representations.  
\end{enumerate}

\subsubsection{Knowledge-Aware Fine-Tuning}
\label{train:ft}

Fine-tuning adapts LLMs to specific domains by training them on relevant datasets, using selected architectures and hyper-parameters to modify the model's weights for improved task performance~\cite{guu2020retrieval, hu2021lora, lu2022dynamic,dettmers2023qlora}. KGs can further tune these models to update and expand their internal knowledge for domain-specific tasks like custom named-entity recognition~\cite{agrawal2023aiseckg}, and text summarization~\cite{kang2022kala}. 

SKILL~\cite{moiseev2022skill} used synthetic sentences converted from WikiData~\cite{seminar2019wikidata} and KELM~\cite{agarwal2020knowledge} used KGs to fine-tune the pre-trained model checkpoints. KGLM~\cite{youn2022kglm} employed an entity-relation embedding layer with KG triples for link prediction tasks. Cross-lingual reasoning~\cite{foroutan2023breaking} improved by fine-tuning MultiLM, mBERT, and mT5 models with logical datasets using a self-attention network. LLMs improve more with additional training using datasets with few-shot CoT reasoning prompts and fine-tuning~\cite{kim2023cot, huang2022large}.

Fine-tuning language models like ChatGPT, limited by their last knowledge update in 2021, is more efficient than training from scratch. It handles queries beyond this cutoff using a curated, domain-specific knowledge graph. The extent to which updated knowledge is integrated into the model remains to be determined. Onoe et al.'s~\cite{onoe2023can} evaluation framework indicate that while models can recall facts about new entities, inferring based on these is harder. The effect of updating knowledge on existing entities is still an open research question.

\subsection{Knowledge-Aware Validation}
\label{sec:KA_valid}

The third category type uses structured data as a fact-checking mechanism and provides a reference for the model to verify information. Knowledge graphs can provide comprehensive explanations and can be used to justify the models' decisions. These methods also help enforce consistency across the facts, obviating the necessity for laborious human-annotated data and enhancing the reliability of generated content. 

The fact-aware language model, KGLM \cite{logan2019barack}, referred to a knowledge graph to generate entities and facts relevant to the context. SURGE~\cite{kang2022knowledge} retrieves high similarity context-relevant triples as a sub-graph from a knowledge graph. 
``Text critic" classifier~\cite{lango2023critic} was proposed to guide the generation by assessing the match between the input data and the generated text.
FOLK~\cite{wang2023explainable} used first-order-logic (FOL) predicates for claim verification in online misinformation. Beyond verification, FOLK generates explicit explanations, providing valuable assistance to human fact-checkers in understanding and interpreting the model's decisions. This approach contributes to the accuracy and interpretability of the model's outputs in the context of misinformation detection.

\begin{table*}[]
\resizebox{\textwidth}{!}{%
\begin{tabular}{llcccc}
\hline
\rowcolor[HTML]{CBCEFB} 
\cellcolor[HTML]{CBCEFB} &
  \cellcolor[HTML]{CBCEFB} &
  \multicolumn{4}{c}{\cellcolor[HTML]{CBCEFB}\textbf{Comparison Attributes}} \\ \cline{3-6} 
\rowcolor[HTML]{CBCEFB} 
\multirow{-2}{*}{\cellcolor[HTML]{CBCEFB}\textbf{Category}} &
  \multirow{-2}{*}{\cellcolor[HTML]{CBCEFB}\textbf{Representative Method}} &
  \textbf{\begin{tabular}[c]{@{}c@{}}Downstream \\ Task\end{tabular}} &
  \textbf{KG Dataset} &
  \textbf{LLM} &
  \textbf{Training} \\ \hline
\rowcolor[HTML]{EFEFEF} 
\cellcolor[HTML]{EFEFEF} &
  \textbf{KAPING \cite{baek2023knowledge}} &
  \textbf{Question-Answering} & 
  \textbf{Mintaka, WebQSP} &
  \textbf{\begin{tabular}[c]{@{}c@{}}T5, T0, OPT, \\ GPT-3\end{tabular}} &
  \cellcolor[HTML]{EFEFEF} \\
\rowcolor[HTML]{EFEFEF} 
\cellcolor[HTML]{EFEFEF} &
  \textbf{Rigel Facts \cite{sen2023knowledge}} &
  \textbf{Question-Answering} &
  \textbf{\begin{tabular}[c]{@{}c@{}}WebQuestions, ComplexWebQuestions, \\ Mintaka, LC-QuAD\end{tabular}} &
  \textbf{\begin{tabular}[c]{@{}c@{}}Flan-T5, T0, \\ OPT, AlexaTM\end{tabular}} &
  \cellcolor[HTML]{EFEFEF} \\
\rowcolor[HTML]{EFEFEF} 
\multirow{-3}{*}{\cellcolor[HTML]{EFEFEF}\textbf{\begin{tabular}[c]{@{}l@{}}KG-\\ Augmented \\ Retrieval\end{tabular}}} &
  \textbf{\begin{tabular}[c]{@{}l@{}}Retrieve-Rewrite-Answer \\ \cite{wu2023retrieve}\end{tabular}} &
  \textbf{Question-Answering} &
  \textbf{MetaQA, WebQSP, WebQ, ZJQA} &
  \textbf{\begin{tabular}[c]{@{}c@{}}ChatGPT, Llama 2, \\ Flan-T5, T0, T5\end{tabular}} &
  \multirow{-3}{*}{\cellcolor[HTML]{EFEFEF}\textbf{\textcircled{X}}} \\
 &
  \textbf{IRCoT \cite{trivedi2022interleaving}} &
  \textbf{Multi-step Reasoning QA} &
  \textbf{\begin{tabular}[c]{@{}c@{}}HotpotQA, 2WikiMultihopQA, \\ MusiQue, IIRC\end{tabular}} &
  \textbf{GPT3, Flan-T5} &
   \\
 &
  \textbf{MindMap \cite{wen2023mindmap}} &
  \textbf{Medical Diagnosis} &
  \textbf{\begin{tabular}[c]{@{}c@{}}GenMedGPT-5k,  \\ CMCQA, ExplainCPE\end{tabular}} &
  \textbf{GPT-3.5, GPT-4} &
   \\
\multirow{-3}{*}{\textbf{\begin{tabular}[c]{@{}l@{}}KG-\\ Augmented \\ Reasoning\end{tabular}}} &
  \textbf{RoG \cite{luo2023reasoning}} &
  \textbf{Reasoning} &
  \textbf{\begin{tabular}[c]{@{}c@{}}WebQSP, \\ Complex WebQuestions (CWQ)\end{tabular}} &
  \textbf{Llama 2-Chat-7B} &
  \multirow{-3}{*}{\textbf{\textcircled{X}}} \\
\rowcolor[HTML]{EFEFEF} 
\cellcolor[HTML]{EFEFEF} &
  \textbf{KnowPrompt \cite{chen2022knowprompt}} &
  \textbf{Relation Extraction and Labeling} &
  \textbf{SemEval, DialogRE, TACRED} &
  \textbf{RoBERTa\_large} &
  \textbf{Few-shot training} \\
\rowcolor[HTML]{EFEFEF} 
\cellcolor[HTML]{EFEFEF} &
  \textbf{BINDER \cite{cheng2022binding}} &
  \textbf{\begin{tabular}[c]{@{}c@{}}Information extraction, \\ Commonsense QA\end{tabular}} &
  \textbf{WikiTableQuestions, TabFact} &
  \textbf{Codex} &
  \textbf{\begin{tabular}[c]{@{}c@{}}API calls / Few-shot \\ In-context learning\end{tabular}} \\
\rowcolor[HTML]{EFEFEF} 
\multirow{-3}{*}{\cellcolor[HTML]{EFEFEF}\textbf{\begin{tabular}[c]{@{}l@{}}Knowledge-\\ Controlled \\ Generation\end{tabular}}} &
  \textbf{BeamQA \cite{atif2023beamqa}} &
  \textbf{Generate Questions} &
  \textbf{MetaQA, WebQSP,} &
  \textbf{T5, BART} &
  \textbf{Fine-tuned for 4 epochs} \\ \hline
 &
  \textbf{SKEP \cite{tian2020skep}} &
  \textbf{Sentiment Analysis} &
  \textbf{\begin{tabular}[c]{@{}c@{}}SST, Amazon, \\ Sem, MPQA\end{tabular}} &
  \textbf{BERT, RoBERTa} &
  \textbf{\begin{tabular}[c]{@{}c@{}}Encoder trained on \\ 3.2m train data\end{tabular}} \\
 &
  \textbf{JointLK \cite{sun2021jointlk}} &
  \textbf{\begin{tabular}[c]{@{}c@{}}Commonsense Question \\ Answering\end{tabular}} &
  \textbf{\begin{tabular}[c]{@{}c@{}}CommonSenseQA, \\ OpenBookQA\end{tabular}} &
  \textbf{RoBERTa-Large} &
  \textbf{\begin{tabular}[c]{@{}c@{}}LM/graph encoder trained \\ jointly for 20 GPU hours\end{tabular}} \\
\multirow{-3}{*}{\textbf{\begin{tabular}[c]{@{}l@{}}Knowledge-\\ Aware \\ Pre-Training\end{tabular}}} &
  \textbf{LKPNR \cite{runfeng2023lkpnr}} &
  \textbf{\begin{tabular}[c]{@{}c@{}}Personalized News \\ Recommendation\end{tabular}} &
  \textbf{MIND} &
  \textbf{\begin{tabular}[c]{@{}c@{}}ChatGLM2, \\ Llama 2, RWKV\end{tabular}} &
  \textbf{\begin{tabular}[c]{@{}c@{}}LK-Encoders trained on \\ GPU for 200K user click logs\end{tabular}} \\
\rowcolor[HTML]{EFEFEF} 
\cellcolor[HTML]{EFEFEF} &
  \textbf{SKILL \cite{moiseev2022skill}} &
  \textbf{Closed-book QA tasks} &
  \textbf{Wikidata, KELM, MetaQA} &
  \textbf{\begin{tabular}[c]{@{}c@{}}T5-base, L, \\ XXL models\end{tabular}} &
  \textbf{T5 fine-tuned for 50k steps} \\
\rowcolor[HTML]{EFEFEF} 
\cellcolor[HTML]{EFEFEF} &
  \textbf{KGLM \cite{youn2022kglm}} &
  \textbf{Link Prediction} &
  \textbf{WN18RR, FB15k-237, UMLS} &
  \textbf{RoBERTa Large} &
  \textbf{Model tuned for 5 epochs} \\
\rowcolor[HTML]{EFEFEF} 
\multirow{-3}{*}{\cellcolor[HTML]{EFEFEF}\textbf{\begin{tabular}[c]{@{}l@{}}Knowledge-\\ Aware \\ Fine-Tuning\end{tabular}}} &
  \textbf{Neurosymbolic \cite{baldazzi2023fine}} &
  \textbf{Banking Customer Query} &
  \textbf{Chase EKG} &
  \textbf{T5-large} &
  \textbf{Model tuned for 10 epochs} \\ \hline
 &
  \textbf{Fact-aware LM \cite{logan2019barack}} &
  \textbf{Fact Generation} &
  \textbf{Linked WikiText-2} &
  \textbf{TransE} &
  \textbf{\begin{tabular}[c]{@{}c@{}}Transformer trained on \\ 256-dim KG embeddings\end{tabular}} \\
 &
  \textbf{SURGE \cite{kang2022knowledge}} &
  \textbf{Dialogue Generation} &
  \textbf{OpenDialKG} &
  \textbf{T5-small} &
  \textbf{\textbf{\textcircled{X}}} \\
\multirow{-3}{*}{\textbf{\begin{tabular}[c]{@{}l@{}}Knowledge-\\ Aware \\ Validation\end{tabular}}} &
  \textbf{FOLK \cite{wang2023explainable}} &
  \textbf{\begin{tabular}[c]{@{}c@{}}Claim Verification in \\ Online Misinformation\end{tabular}} &
  \textbf{\begin{tabular}[c]{@{}c@{}}HoVER, FEVEROUS,\\  SciFact-Open\end{tabular}} &
  \textbf{\begin{tabular}[c]{@{}c@{}}Llama(7B), Llama(13B), \\ Llama(30B)\end{tabular}} &
  \textbf{\textcircled{X}} \\ \hline
\end{tabular}%
}
\caption{Comparison attributes of Knowledge Graph-enhanced LLM methods}
\label{tab:comp}
\end{table*}

\section{Discussion, Challenges and Future}
\label{discussion}
In this section, we examine the effectiveness of KG-enhanced LLM techniques in reducing hallucinations and enhancing performance and reliability in LLMs. We also identify key challenges associated with each method and propose potential research avenues in this evolving field.

\subsection{Resources}
Table~\ref{tab:comp} details the key features of different KG-enhanced LLM methods, emphasizing their application in specific industries using domain-specific knowledge graphs. The inference methods used general knowledge and commonsense reasoning datasets for QA tasks without requiring LLM retraining. Mindmap~\cite{wen2023mindmap} demonstrated an application in healthcare, augmenting clinical datasets with GPT-4. Meng et al.~\cite{meng2021rewire} pre-trained T5 and BART models using a biomedical knowledge graph, Unified Medical Language System (UMLS) Metathesaurus. LKPNR~\cite{runfeng2023lkpnr} pre-trained LM and graph encoders on MIND-200K user click logs to provide personalized news recommendations. 
Martino et al.~\cite{martino2023knowledge} used knowledge injection to reduce hallucinations when responding to online customer reviews for a retail store. 
Dong et al.~\cite{dong2022faithful} showed improvement in the faithfulness of text summarization tasks to the source documents by linking external source knowledge bases from the source. Baldazzi~\cite{baldazzi2023fine} fine-tuned T5-large on financial customer-service enterprise KG. 

\subsection{Evaluation Metrics}
Various criteria were applied to assess the effectiveness of knowledge graph augmentation in reducing hallucinations in LLMs. 

\noindent \textit{\textbf{Accuracy:}} Accuracy comparison with and without augmented knowledge from KGs~\cite{baek2023knowledge, zhang2023iag}.
\\
\noindent \textit{\textbf{Top-K and MRR:}} 
Retrieval performance was measured by the relevance of retrieved triples for generating answers. Mean Reciprocal Rank (MRR) and Top-K accuracy determined the ranks of correctly retrieved answer-containing triples~\cite{baek2023knowledge, sen2023knowledge}. The effectiveness of KG triples was assessed as either "Helpful" or "Harmful" and compared against scenarios where "no knowledge" was provided~\cite{wu2023retrieve}. 
\\
\noindent \textit{\textbf{Hits@1:}} Evaluates answer accuracy and examines the coverage of multi-choice question answers~\cite{luo2023reasoning, wu2023retrieve,wei2023kicgpt}. 
\\
\noindent \textit{\textbf{Execution Accuracy (EA):}} The controlled generation method, such as Binder~\cite{cheng2022binding}, uses Execution Accuracy (EA) as a metrics to measure the accuracy in semantic parsing, API call generation, and the success rate of code execution.
\\
\noindent \textit{\textbf{Exact Match (EM):}} Model's performance after fine-tuning was evaluated using EM (Exact Match) scores on test sets~\cite{moiseev2022skill}.
\\
\noindent \textit{\textbf{Human Evaluation:}} Validation methods were manually evaluated to assess the explanation quality, coverage, logical soundness, fluency, and factual accuracy of sentence completion~\cite{wang2023explainable, kang2022knowledge}.
It is pertinent to consider evaluating factuality from different aspects, first verifying the presence of accurate and reliable information and second identifying any instances of fabricated or "hallucinated" information.

\subsection{Performance Analysis}
\textbf{Retrieved facts enhance small LLMs:} 
Smaller models, due to their limited parameter spaces, struggle to incorporate extensive knowledge in pre-training. Augmenting facts from knowledge graphs, rather than increasing model size, enhanced answer correctness by over $80\%$ for question-answering tasks~\cite{baek2023knowledge,sen2023knowledge,wu2023retrieve}. 
However, the success of these methods with complex queries heavily relies on the retriever modules, whose capabilities are limited to the knowledge graph~\cite{behnamghader2022can}.

\noindent \textbf{Step-wise reasoning more effective in larger models:} 
Variations of CoT methods offer cost-effective control and task-specific tuning, enhancing model performance. For instance, RoG~\cite{luo2023reasoning} reported an increase in ChatGPT's accuracy from $66.8\%$ to $85.7\%$ in reasoning tasks with knowledge graph augmentation. Similarly, Mindmap~\cite{wen2023mindmap} boosted accuracy in disease diagnosis and drug recommendation to $88.2\%$ using a clinical reasoning graph.

\noindent \textbf{Controlled generation boosts the performance:} Knowledge-controlled generation methods surpass baseline models in accuracy and contextual relevance, enhancing their ability to handle diverse queries~\cite{chen2022knowprompt, cheng2022binding, atif2023beamqa}. However, these methods can vary in quality and are sometimes prone to generating incorrect or irrelevant information.

\noindent \textbf{Pre-training and fine-tuning are costly:} Pre-training and fine-tuning significantly enhance domain-specific task performance. However, these improvements require substantial computational resources, as shown in Table~\ref{tab:comp}. Additionally, fine-tuning's data-dependency makes it task-specific and limits its transferability and generalizability~\cite{gueta2023knowledge, wang2023explainable}.

\noindent \textbf{Fact-checking ensures reliability:} Knowledge validation through fact-checking reduces hallucinations by checking model-generated data against a knowledge graph, but it increases computational load and may miss some inaccuracies~\cite{kang2022knowledge,lango2023critic}. 

The effectiveness of knowledge augmentation is also influenced by the size of the knowledge graph and its impact on query responses. Standard approaches include fine-tuning pre-trained models for reliability but at a higher cost, and example-based prompting, less effective in certain reasoning tasks~\cite{brown2020language,rae2021scaling}. Zhang et al.~\cite{zhang2023language} noted that language model inconsistencies often arise from incorrect context usage. Method selection depends on the specific use case and available resources. Wang et al.~\cite{wang2023shall} showed that pre-training decoder-only LLMs with retrieval can improve factual accuracy in knowledge-intensive tasks, while Shi et al.~\cite{shi2023hallucination} developed GraphNarrative, a dataset aimed at reducing hallucinations, beneficial for fine-tuning LLMs.

\subsection{Trend Analysis}
\begin{figure}
    \centering
    \includegraphics[width=0.48\textwidth]{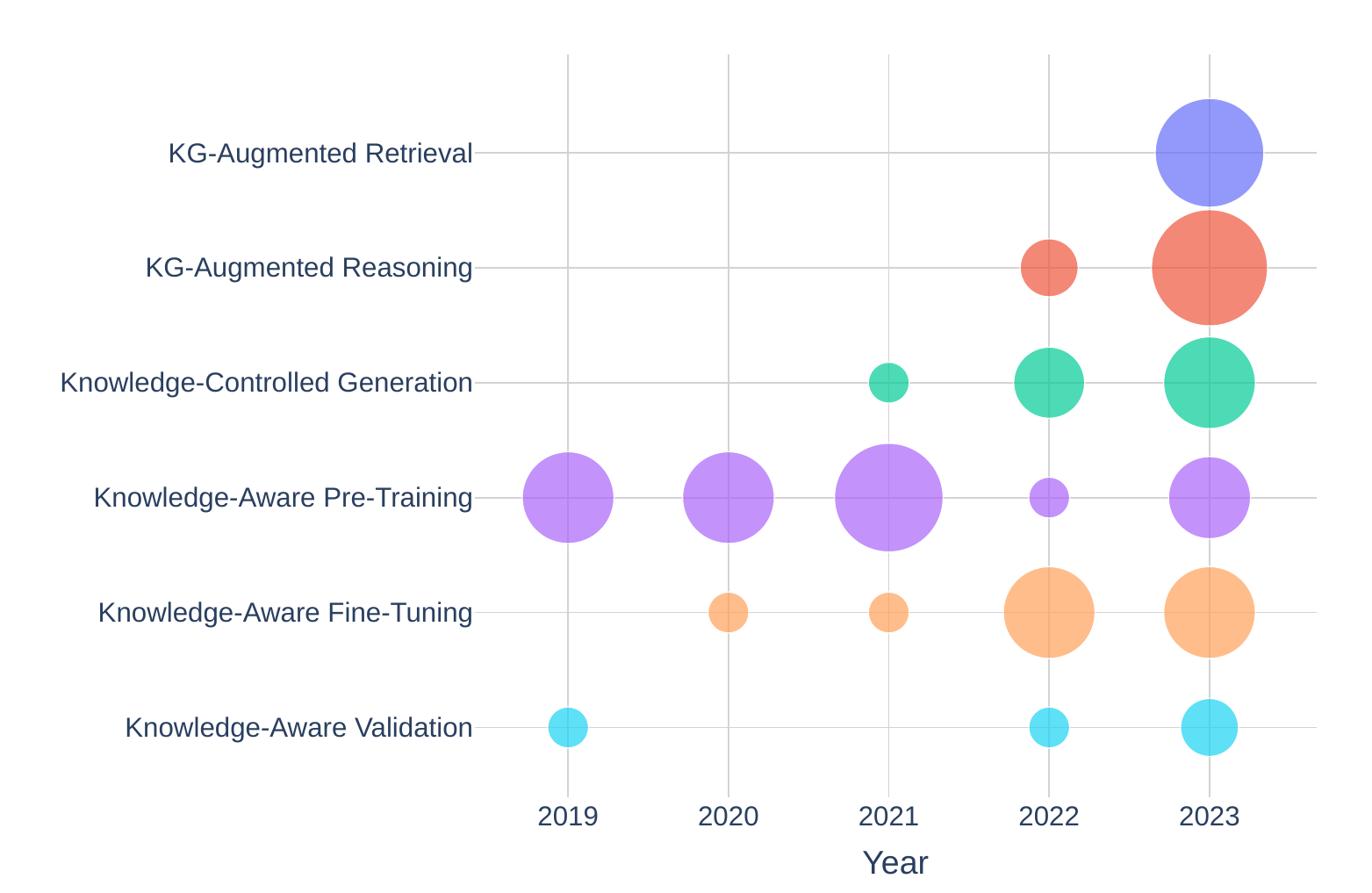}
    \caption{Research trend over years- The bubble size represents number of papers we observed for each knowledge-graph augmentation categories: smallest size (\#papers=1), largest size (\#papers=8)}
    \label{fig:trend}
\end{figure}

\textbf{Figure} \ref{fig:trend} shows the research trends using different knowledge-graph augmentation techniques from 2019 to 2023. The bubble size here represents the number of papers for each knowledge-graph augmentation category, ranging from one to eight. Pre-training methods by adding knowledge graphs to the training corpus were predominant in the early years of language model development. After the extensive GPT series of LLMs, retraining the huge model with billions of parameters became impractical and resource-intensive. More efforts were made to fine-tune the models with task-specific data without training from scratch. Very recently, there has been a shift towards using knowledge-augmented retrieval, reasoning, generation, and validation methods without incurring additional training costs.

\subsection{Future Directions}
\label{future}
Here are some potential future research directions for further investigation:

\noindent \textbf{Improve Quality of KG:} \textbf{\textcircled{a}Context-Aware:} Dynamic KGs that continuously adapt to changing contexts and new information can improve LLMs effectively. \textbf{\textcircled{b}Addressing Biases:}Fairness-aware algorithms in KGs can ensure bias or misinformation is not perpetuated by KGs. \textbf{\textcircled{c}Cross-Domain Knowledge:} Integrating knowledge from diverse domains like science, art, and history into a single graph could enhance the depth and nuance of LLM responses.\textbf{\textcircled{d}Multi-Modal:} Adding multi-modal data such as images, videos, and audio to KGs can enrich the data pool and improve LLMs' contextual responses.
\\
\\
\noindent \textbf{Mixture of Experts (MoE) LLMs:} Efforts are underway to optimize the MoE architecture to scale LLMs and increase their capacity without increasing computation~\cite{zhou2022mixture}. Integrating MoE with knowledge graphs~\cite{yu2022diversifying} can develop adaptive learning strategies for context-based expert utilization and improve the interpretability and transparency of MoE-LLMs.
\\
\\
\noindent \textbf{Symbolic-Subsymbolic Unification:} Knowledge fabrics, such as symbolic KGs and sub-symbolic vectors, enables versatile reasoning in LLMs, mimicking human mind's capacity to reconcile structured theories~\cite{nunez2023review}.
\\
\\
\noindent \textbf{Synergizing LLM and KG:} 
LLMs are being used for link prediction and knowledge graph completion~\cite{xiao2023instructed, veseli2023evaluating}. Synergizing the LLM and KGs is a potential direction where both components can mutually enhance each other's capabilities through a bidirectional reasoning process driven by a harmonious blend of data and knowledge~\cite{pan2023unifying}.
\\
\\
\noindent \textbf{Causality-Awareness:} Incorporating causality into knowledge graphs, ~\cite{wei2022causal}, will enhance Large Language Models' (LLMs) capability to grasp causation rather than merely identifying correlations. This advancement will equip LLMs with a better understanding of the causal relationships between events or entities, significantly improving their reasoning and predictive capabilities.

The progress of KGs promises to greatly enhance LLMs, making them more relevant, responsive, and accurate. This aims to create more reliable and trustworthy language models, advancing robust and responsible AI systems.

\section{Conclusion}
In this survey, we systematically investigate the integration of KGs into LLMs to mitigate hallucinations and improve reasoning accuracy. We emphasize the benefits of using KGs to enhance LLM performance across various phases at inference, model training, and output verification stages. While substantial progress has been made, we emphasize the need for continuous innovation and propose future directions to facilitate the development of more advanced KG-augmented LLMs.

\section{Limitations}
In this paper, we conduct a comprehensive review of knowledge-graph-based augmentation techniques in LLMs, with a specific focus on their ability to address hallucinations. We identify commonalities among these techniques and categorize them into three distinct groups based on their mechanisms and approaches. Furthermore, we systematically assess the performance of these methods. In Section \ref{intro}, we compare our work with existing related surveys and we will continue adding more related approaches. However, it's important to acknowledge that despite our diligent efforts, there may be certain limitations that still exist in this paper.
\\
\\
\noindent \textbf{References and Methods.} Due to page limitations, we may not include all relevant references and detailed technical information. Our study primarily focuses on state-of-the-art methods developed between 2019 and 2023, sourced primarily from reputable conferences and platforms such as ACL, EMNLP, NAACL, ICLR, ICML, and arXiv. We remain committed to keeping our work up-to-date.
\\
\\
\noindent \textbf{Taxonomy and Comparison.} We primarily categorized the methods based on their primary augmentation approach. In some cases, hybrid studies incorporating multiple approaches may be categorized differently, depending on specific criteria. It's essential to note that our analysis is based on the performance of existing works using the current experiments and datasets. Given the rapid evolution in this field, benchmarks and baseline models may change, potentially leading to variations in these evaluations.

\section*{Acknowledgements}
This material is based upon work supported by the National Science Foundation under Grant No.  2114789.

\bibliography{anthology,ref}

% \appendix

% \section{Example Appendix}
% \label{sec:appendix}

% This is an appendix.

\end{document}